\documentclass[10pt,twocolumn,letterpaper]{article}
 
\usepackage{cvpr}
\usepackage{times}
\usepackage{epsfig}
\usepackage{graphicx}
\usepackage{amsmath}
\usepackage{amssymb}
\usepackage{caption}
\usepackage{booktabs}

\newcommand{\figref}[1]{Fig.~\ref{#1}}

\def\fps{-fps\@\xspace}

\usepackage[pagebackref=true,breaklinks=true,letterpaper=true,colorlinks,citecolor=blue,linkcolor=blue,bookmarks=false]{hyperref}

\cvprfinalcopy 

\def\cvprPaperID{325} 

\ifcvprfinal\fi
\begin{document}

\title{Super SloMo: High Quality Estimation of Multiple Intermediate Frames \\for  Video Interpolation}


\author{
Huaizu Jiang$^1$~~~~~~~~Deqing Sun$^2$~~~~~~~~Varun Jampani$^2$\\
Ming-Hsuan Yang$^{3,2}$~~~~~~~~Erik Learned-Miller$^1$~~~~~~~~Jan Kautz$^2$\\
$^1$UMass Amherst~~~~~$^2$NVIDIA~~~~~$^3$UC Merced\\
{\tt\small \{hzjiang,elm\}@cs.umass.edu,\{deqings,vjampani,jkautz\}@nvidia.com, mhyang@ucmerced.edu}
}

\maketitle

\begin{abstract}
Given two consecutive frames, video interpolation aims at generating intermediate frame(s) to form both spatially and temporally coherent video sequences. While most existing methods focus on \emph{single-frame} interpolation, we propose an end-to-end convolutional neural network for \emph{variable-length multi-frame} video interpolation, where the motion interpretation and occlusion reasoning are jointly modeled. We start by computing  bi-directional optical flow between the input images using a U-Net architecture. These flows are then linearly combined at each time step to approximate the intermediate bi-directional optical flows. These approximate flows, however, only work well in locally smooth regions and produce artifacts around motion boundaries. To address this shortcoming, we employ another U-Net to refine the approximated flow and also predict soft visibility maps.  
Finally, the two input images are warped and linearly fused to form each intermediate frame. By applying the visibility maps to the warped images before fusion, we exclude the contribution of occluded pixels to the interpolated intermediate frame to avoid artifacts. Since none of our learned network parameters are time-dependent, our approach is able to produce as many intermediate frames as needed. To train our network, we use 1,132 240\fps video clips, containing 300K individual video frames. Experimental results on several datasets, predicting different numbers of interpolated frames, demonstrate that our approach performs consistently better than existing methods.
\end{abstract}


\section{Introduction}
There are many memorable moments in your life that you might want to record with a camera in \emph{slow-motion} because they are hard to see clearly with your eyes: the first time a baby walks, a difficult skateboard trick, a dog catching a ball, \emph{etc}. While it is possible to take 240\fps (frame-per-second) videos with a cell phone, professional high-speed cameras are still required for higher frame rates. In addition, many of the moments we would like to slow down are unpredictable, and as a result, are recorded at standard frame rates. Recording everything at high frame rates is impractical--it requires large memories and is power-intensive for mobile devices. 

Thus it is of great interest to generate high-quality slow-motion video from existing videos. In addition to transforming standard videos to higher frame rates, video interpolation can be used to generate smooth view transitions. It also has intriguing new applications in self-supervised learning, serving as a supervisory signal to learn optical flow from unlabeled videos~\cite{liu17video,long16learning}.

It is challenging to generate multiple intermediate video frames because the frames have to be coherent, both spatially and temporally. For instance, generating 240\fps videos from standard sequences (30-fps) requires interpolating seven intermediate frames for every two consecutive frames. A successful solution has to not only correctly interpret the motion between two input images (implicitly or explicitly), but also understand occlusions. Otherwise, it may result in severe artifacts  in the interpolated frames, especially around motion boundaries. 

Existing methods mainly focus on \emph{single-frame} video interpolation and have achieved impressive performance for this problem setup~\cite{liu17video,long16learning,niklaus17video_cvpr,niklaus17video_iccv}. However, these methods cannot be directly used to generate arbitrary higher frame-rate videos. While it is an appealing idea to apply a single-frame video interpolation method recursively to generate multiple intermediate frames, this approach has at least two limitations. First, recursive single-frame interpolation cannot be fully parallelized, and is therefore slow, since some frames cannot be computed until other frames are finished (e.g., in seven-frame interpolation, frame~2 depends on 0 and 4, while frame~4 depends on 0 and 8).
 Errors also accumulates during recursive interpolation. Second, it can only generate $2^i\!-\!1$ intermediate frames (\eg, 3, 7). As a result, one cannot use this approach (efficiently) to generate 1008\fps video from 24\fps, which requires generating 41 intermediate frames. 

In this paper we present a high-quality \emph{variable-length multi-frame} interpolation method that can interpolate a frame at any arbitrary time step between two frames.  Our main idea is to warp the input two images to the specific time step and then adaptively fuse the two warped images to generate the intermediate image, where the motion interpretation and occlusion reasoning are modeled in a single end-to-end trainable network. Specifically, we first use a \emph{flow computation} CNN to estimate the bi-directional optical flow between the two input images, which is then linearly fused to approximate the required intermediate optical flow in order to warp input images. This approximation works well in smooth regions but poorly around motion boundaries. We therefore use another \emph{flow interpolation} CNN to refine the flow approximations and predict soft visibility maps. By applying the visibility maps to the warped images before fusion, we exclude the contribution of occluded pixels to the interpolated intermediate frame, reducing artifacts. 
The parameters of both our flow computation and interpolation networks are independent of the specific time step being interpolated, which is an input to the flow interpolation network.
Thus, our approach can generate as many intermediate frames as needed in parallel, 


To train our network, we collect 240\fps videos from YouTube and hand-held cameras~\cite{su16deep}.
In total, we have 1.1K video clips, consisting of 300K individual video frames with a typical resolution of $1080\times 720$. We then evaluate our trained model on several other independent datasets that  require different numbers of interpolations, including the Middlebury~\cite{baker11a}, UCF101~\cite{ucf101},  slowflow dataset~\cite{Janai2017CVPR}, and high-frame-rate MPI Sintel~\cite{Janai2017CVPR}. Experimental results demonstrate that our approach significantly outperforms existing methods on all datasets. We also evaluate our unsupervised (self-supervised) optical flow results on the KITTI 2012 optical flow benchmark~\cite{Geiger:2012:KITTI} and obtain better results than the recent method~\cite{liu17video}.

\section{Related Work}
\noindent\textbf{Video interpolation.} 
The classical approach to video interpolation is based on optical flow~\cite{Baker2009OcclusionInterpolation,Barron:1994:PO}, and interpolation accuracy is often used to evaluate optical flow algorithms~\cite{baker11a,szeliski1999prediction}. Such approaches can generate intermediate frames at arbitrary times between two input frames. 
Our experiments show that state-of-the-art optical flow method~\cite{ilg16flownet2}, coupled with occlusion reasoning~\cite{baker11a}, can serve as a strong baseline for frame interpolation. However, motion boundaries and severe occlusions are still challenging to existing flow methods~\cite{Butler:ECCV:2012,Geiger:2012:KITTI}, and thus the interpolated frames tend to have artifacts around boundaries of moving objects. Furthermore, the intermediate flow computation (\ie, flow interpolation) and occlusion reasoning are based on heuristics and not end-to-end trainable. 

Mahajan \etal~\cite{mahajan2009moving}  move the image gradients to a given time step and solve a Poisson equation to reconstruct the interpolated frame. This method can also generate multiple intermediate frames, but is computationally expensive because of the complex optimization problems.
Meyer \etal~\cite{meyer15phase}  propose propagating phase information across oriented multi-scale pyramid levels for video interpolation. While achieving impressive performance, this method still tends to fail for high-frequency contents with large motions.

The success of deep learning in high-level vision tasks has inspired numerous deep models for low-level vision tasks, including frame interpolation. Long \etal~\cite{long16learning} use frame interpolation as a supervision signal to learn CNN models for optical flow. However, their main target is optical flow and the interpolated frames tend to be blurry. Niklaus \etal~\cite{niklaus17video_cvpr} consider the frame interpolation as a local convolution over the two input frames and use a CNN to learn a spatially-adaptive convolution kernel for each pixel. Their method obtains high-quality results.
However, it is both computationally expensive and memory intensive to predict a kernel for every pixel. Niklaus \etal~\cite{niklaus17video_iccv} improve the efficiency by predicting separable kernels. But the motion that can be handled is limited by the kernel size (up to 51 pixels). Liu \etal~\cite{liu17video} develop a CNN model for frame interpolation that has an explicit sub-network for motion estimation. Their method obtains not only good interpolation results but also promising unsupervised flow estimation results on KITTI 2012. 
However, as discussed previously, these CNN-based single-frame interpolation methods~\cite{niklaus17video_cvpr,niklaus17video_iccv,liu17video} are not well-suited for multi-frame interpolation. 

Wang~\etal~\cite{wang17light} investigate to generate intermediate frames for a light field video using video frames taken from another standard camera as references. In contrast, our method aims at producing intermediate frames for a plain video and does not need reference images.
 





\noindent\textbf{Learning optical flow.} 
State-of-the-art optical flow methods~\cite{Wulff2017Optical,Xu2017Accurate} adopt the variational approach introduce by Horn and Schunck~\cite{Horn:1981:DO}. Feature matching is often adopted to deal with small and fast-moving objects~\cite{Brox:LDOF:2011,EpicFlow}.  However, this approach requires the optimization of a complex objective function and is often computationally expensive. Learning is often limited to a few parameters~\cite{Li2008Learning,Roth:2007:SSO,Sun:2008:LOF}.

Recently, CNN-based models are becoming increasingly popular for learning optical flow between input images.
Dosovitskiy \etal\cite{Dosovitskiy:2015Flownet} develop two network architectures, FlowNetS and FlowNetC, and show the feasibility of learning the mapping from two input images to optical flow using CNN models. Ilg \etal~\cite{ilg16flownet2} further use the FlowNetS and FlowNetC as building blocks to design a larger network, FlowNet2,  to achieve much better performance. Two recent methods have also been proposed~\cite{Ranjan:2016:SpyNet,sun2017pwc} to build the classical principles of optical flow into the network architecture, achieving comparable or even better results and requiring less computation than FlowNet2~\cite{ilg16flownet2}.  

In addition to the supervised setting, learning optical flow using CNNs in an unsupervised way has also been explored. The main idea is to use the predicted flow to warp one of the input images to another. The reconstruction error serves as a supervision signal to train the network. Instead of merely considering two frames~\cite{yu16back}, a memory module is proposed to keep the temporal information of a video sequence~\cite{patraucean16spatio}. Similar to our work, Liang \etal~\cite{liang17dual} train optical flow via video frame extrapolation, but their training uses the flow estimated by the EpicFlow method~\cite{EpicFlow} as an additional supervision signal. 



\section{Proposed Approach}
In this section, we first introduce optical flow-based intermediate frame synthesis in section~\ref{subsec:frame_synth}. We then explain details of our flow computation and flow interpolation networks in section~\ref{subsec:flow_interp}. In section~\ref{subsec:training}, we define the loss function used to train our networks.

\subsection{Intermediate Frame Synthesis}
\label{subsec:frame_synth}
Given two input images $I_0$ and $I_1$ and a time $t\in(0, 1)$, our goal is to predict the intermediate image $\hat{I}_t$ at time $T=t$. 
A straightforward way is to accomplish this is to train a neural network~\cite{long16learning} to directly output the RGB pixels of $\hat{I}_t$. 
In order to do this, however, the network has to learn to interpret not only the motion pattens but also the appearance of the two input images. 
Due to the rich RGB color space, it is hard to generate high-quality intermediate images in this way. 
Inspired by~\cite{baker11a} and recent advances in single intermediate video frame interpolation~\cite{niklaus17video_cvpr,niklaus17video_iccv,liu17video}, we propose fusing the warped input images at time $T=t$. 

Let $F_{t\rightarrow0}$ and $F_{t\rightarrow1}$ denote the optical flow from $I_t$ to $I_0$ and $I_t$ to $I_1$, respectively. If these two flow fields are known, we can synthesize the intermediate image $\hat{I}_t$ as follows:
\begin{align}
\hat{I}_t = \alpha_0\odot g(I_0, F_{t\rightarrow0}) + (1 - \alpha_0)\odot g(I_1, F_{t\rightarrow1}),
\end{align}
where $g(\cdot, \cdot)$ is a \emph{backward warping} function, which can be implemented using bilinear interpolation~\cite{zhou16view,liu17video} and is differentiable. The parameter $\alpha_0$ controls the contribution of the two input images and depend on two factors: temporal consistency and occlusion reasoning. $\odot$ denotes element-wise multiplication, implying content-aware weighting of input images. For temporal consistency, the closer the time step $T=t$ is to $T=0$, the more contribution $I_0$ makes to $\hat{I}_t$; a similar property holds for $I_1$. On the other hand, an important property of the video frame interpolation problem is that if a pixel $p$ is visible at $T=t$, it is most likely \emph{at least visible in one of the input images},\footnote{It is a rare case but it may happen that an object appears and disappears between $I_0$ and $I_1$.} which means the occlusion problem can be addressed.
We therefore introduce \emph{visibility maps} $V_{t\leftarrow0}$ and $V_{t\leftarrow1}$. $V_{t\leftarrow0}(p)\in[0,1]$ denotes whether the pixel $p$ remains visible (0 is fully occluded) when moving from $T=0$ to $T=t$. Combining the temporal consistency and occlusion reasoning, we have
\begin{align}
\hat{I}_t \!=\! \frac{1}{Z}\!\odot\!\big((1\!-\!t)V_{t\leftarrow0}\!\odot \!g(I_0, F_{t\rightarrow0}) \!+\! tV_{t\leftarrow1}\!\odot \!g(I_1, F_{t\rightarrow1})\big)\notag,
\end{align}
where $Z=(1-t)V_{t\rightarrow 0}+tV_{t\rightarrow 1}$ is a normalization factor.

\subsection{Arbitrary-time Flow Interpolation}
\label{subsec:flow_interp}
Since we have no access to the target intermediate image $I_t$, it is hard to compute the flow fields $F_{t\rightarrow0}$ and $F_{t\rightarrow1}$. To address this issue, we can approximately synthesize the intermediate optical flow $F_{t\rightarrow0}$ and $F_{t\rightarrow1}$ using the optical flow between the two input images $F_{0\rightarrow1}$ and $F_{1\rightarrow0}$.

\begin{figure}[t]
\centering
\includegraphics[width=0.6\linewidth]{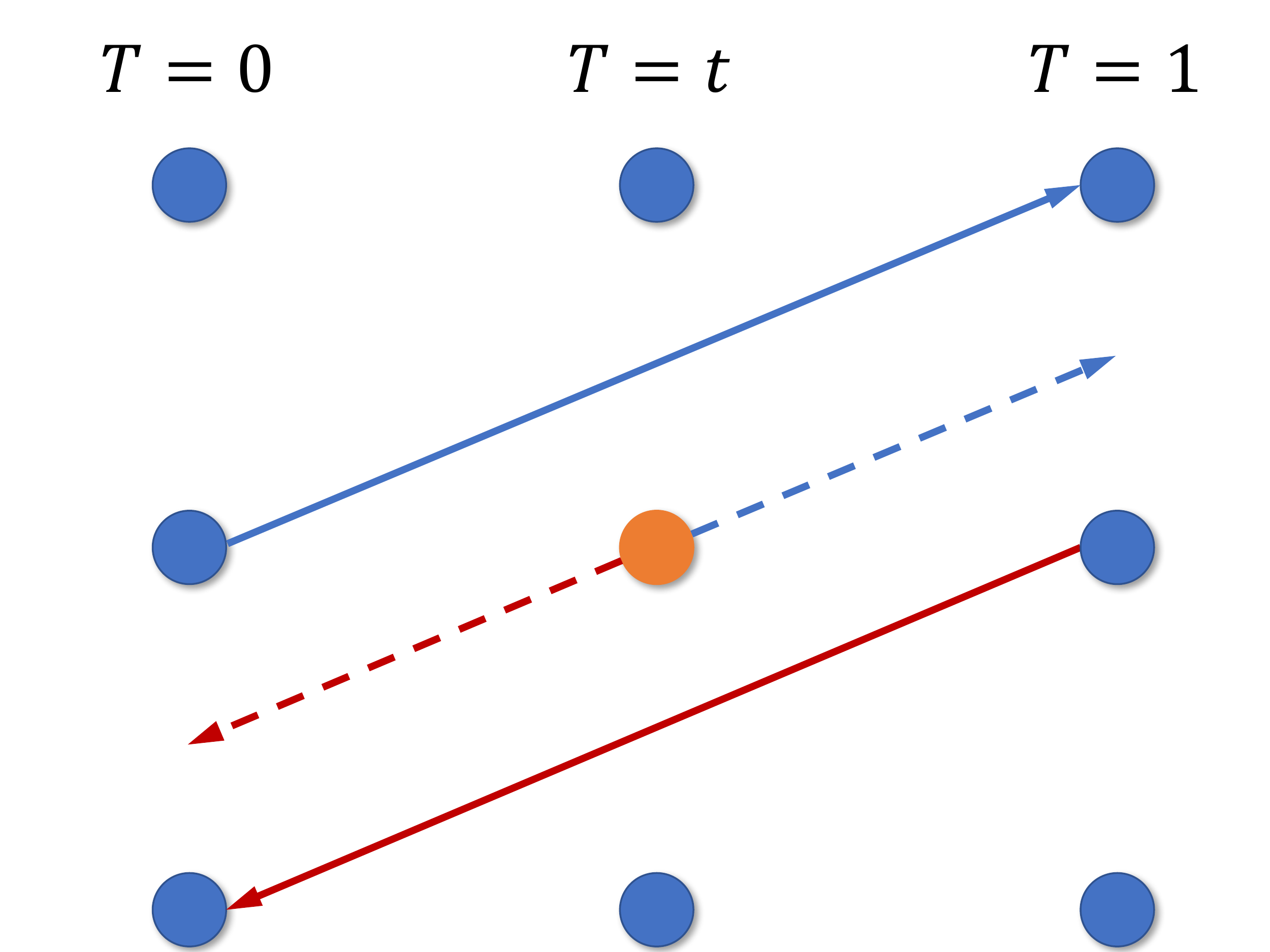}
\caption{Illustration of intermediate optical flow approximation. The orange pixel borrows optical flow from pixels at the same position in the first and second images.}
\label{fig:flow_approx_illustration}
\end{figure}

Consider the toy example shown in~\figref{fig:flow_approx_illustration}, where each column corresponds to a certain time step and each dot represents a pixel. For the orange dot $p$ at $T=t$, we are interested in synthesizing its optical flow to its corresponding pixel at $T=1$ (the blue dashed arrow). One simple way is to borrow the optical flow \emph{from the same grid positions} at $T=0$ and $T=1$ (blue and red solid arrows), assuming that the optical flow field is locally smooth. Specifically, $F_{t\rightarrow1}(p)$ can be approximated as
\begin{align}
\hat{F}_{t\rightarrow1}(p) &= (1-t)F_{0\rightarrow1}(p)\\
&\mbox{or}\notag\\
\hat{F}_{t\rightarrow1}(p) &= -(1-t)F_{1\rightarrow0}(p),
\label{eq:flow_approx}
\end{align}
where we take the direction of the optical flow between the two input images in the same or opposite directions and scale the magnitude accordingly ($(1-t)$ in~\eqref{eq:flow_approx}). Similar to the temporal consistency for RGB image synthesis, we can approximate the intermediate optical flow by combining the bi-directional input optical flow as follows (in vector form).
\begin{align}
\hat{F}_{t\rightarrow 0} &= -(1-t)tF_{0\rightarrow 1} + t^2F_{1\rightarrow 0} \notag\\
\hat{F}_{t\rightarrow 1} &= (1-t)^2F_{0\rightarrow 1} -t(1-t)F_{1\rightarrow 0}.
\label{eq:flow_approx_bidirect}
\end{align}

\begin{figure}[t]
\centering
\renewcommand{\tabcolsep}{1pt}
\begin{tabular}{ccc}
\includegraphics[width=0.33\linewidth]{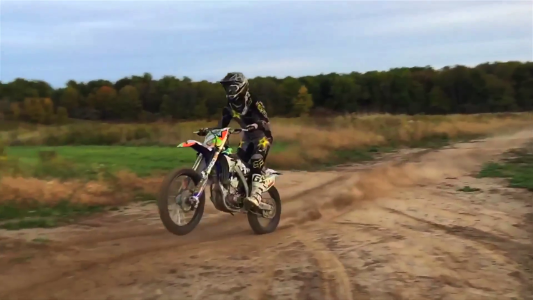} & 
\includegraphics[width=0.33\linewidth]{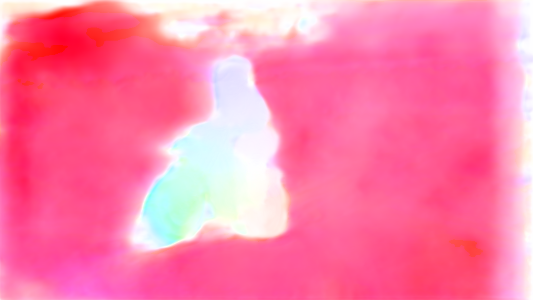} & 
\includegraphics[width=0.33\linewidth]{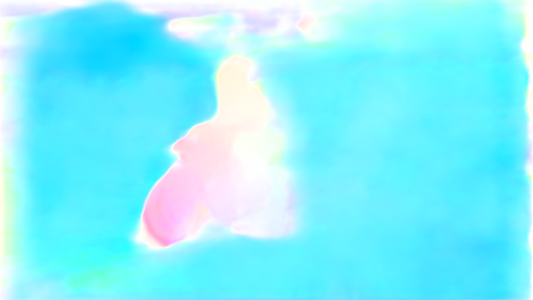} \\
$I_0$ & $F_{0\rightarrow1}$ & $F_{1\rightarrow0}$ \\
\includegraphics[width=0.33\linewidth]{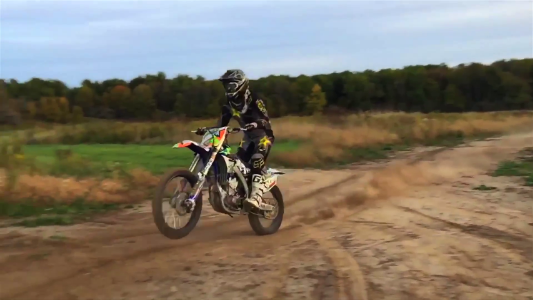} & 
\includegraphics[width=0.33\linewidth]{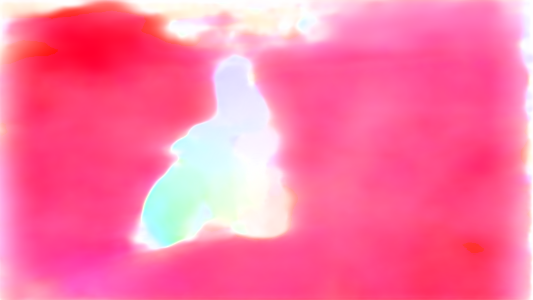} & 
\includegraphics[width=0.33\linewidth]{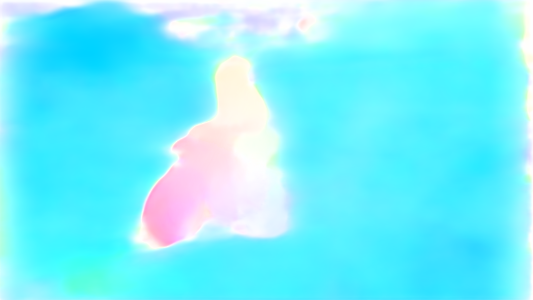} \\
$I_t$ & $\hat{F}_{t\rightarrow1}$ & $\hat{F}_{t\rightarrow0}$ \\
\includegraphics[width=0.33\linewidth]{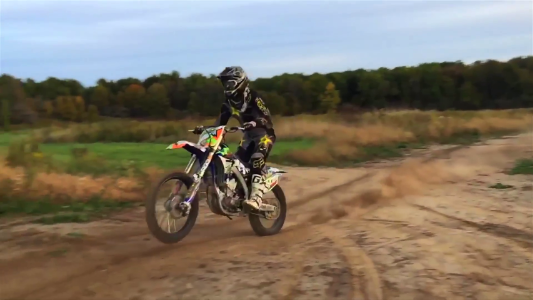} & 
\includegraphics[width=0.33\linewidth]{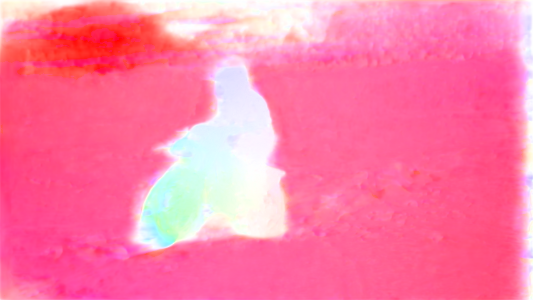} & 
\includegraphics[width=0.33\linewidth]{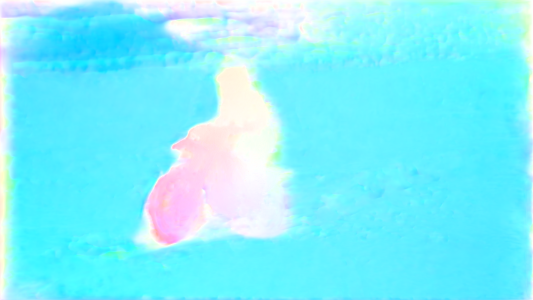} \\
$I_1$ & $F_{t\rightarrow1}$ & $F_{t\rightarrow0}$ \\
 & 
\includegraphics[width=0.33\linewidth]{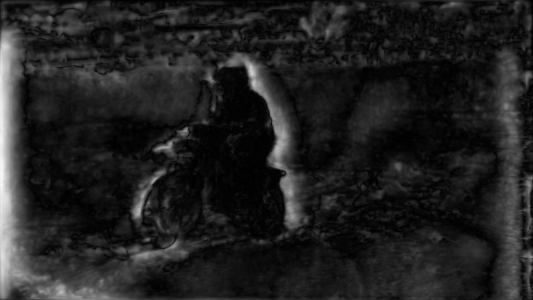} & 
\includegraphics[width=0.33\linewidth]{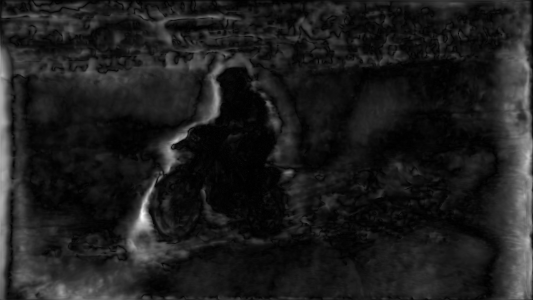} 
\\
 & $\|F_{t\rightarrow1}-\hat{F}_{t\rightarrow1}\|_2$ & $
 \|F_{t\rightarrow0}-\hat{F}_{t\rightarrow0}
 \|_2$ \\
\end{tabular}
\caption{Samples of flow interpolation results, where $t=0.5$. The entire scene is moving toward the left (due to camera translation) and the motorcyclist is independently moving left. The last row shows that the refinement from our flow interpolation CNN is mainly around the motion boundaries (the whiter a pixel, the bigger the refinement).}
\label{fig:flow_interpolation}
\end{figure}

This approximation works well in smooth regions but poorly around motion boundaries, because the motion near motion boundaries is not locally smooth.  To reduce artifacts around motion boundaries, which may cause poor image synthesis, we propose learning to refine the initial approximation. Inspired by the cascaded architecture for optical flow estimation in~\cite{ilg16flownet2}, we train a flow interpolation sub-network. This sub-network takes the input images $I_0$ and $I_1$, the optical flows between them $F_{0\rightarrow1}$ and $F_{0\rightarrow1}$, the flow approximations $\hat{F}_{t\rightarrow 0}$ and $\hat{F}_{0\rightarrow 1}$, and two warped input images using the approximated flows $g(I_0, \hat{F}_{t\rightarrow0})$ and $g(I_1, \hat{F}_{t\rightarrow1})$ as input, and outputs refined intermediate optical flow fields $F_{t\rightarrow1}$ and $F_{t\rightarrow0}$. Sample interpolation results are displayed in Figure~\ref{fig:flow_interpolation}.

\begin{figure}
\centering
\setlength{\belowcaptionskip}{-10pt}
\renewcommand{\tabcolsep}{1pt}
\begin{tabular}{cc}
\includegraphics[width=0.43\linewidth]{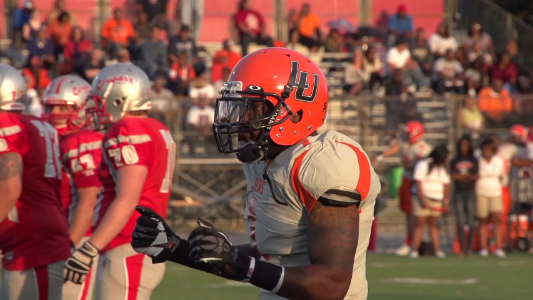} & 
\includegraphics[width=0.43\linewidth]{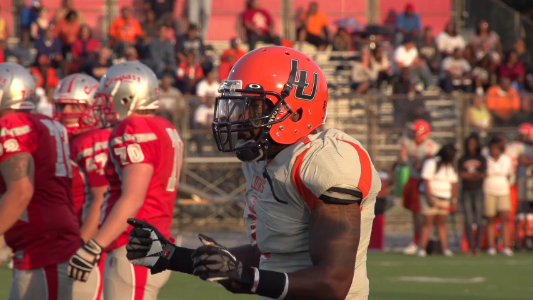} \\
$I_0$ & $I_1$ \\
\includegraphics[width=0.43\linewidth]{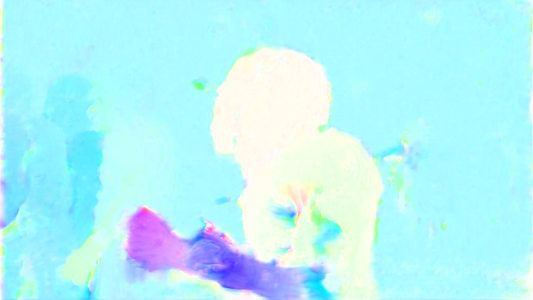} & 
\includegraphics[width=0.43\linewidth]{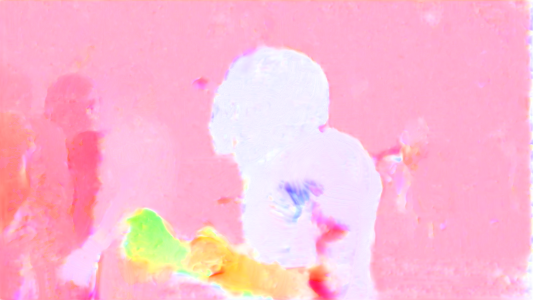} \\
$F_{t\rightarrow0}$ & $F_{t\rightarrow1}$ \\
\includegraphics[width=0.43\linewidth]{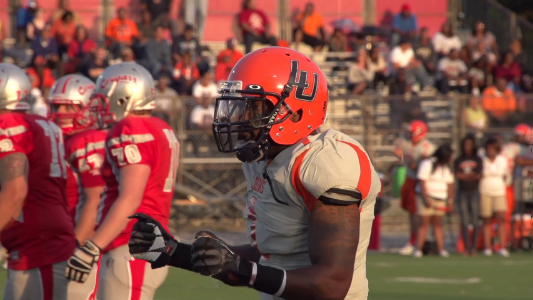} & 
\includegraphics[width=0.43\linewidth]{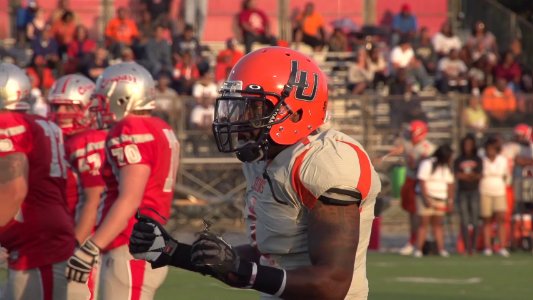} \\
$g(I_0, F_{t\rightarrow0})$ & $g(I_1, F_{t\rightarrow1})$ \\
\includegraphics[width=0.43\linewidth]{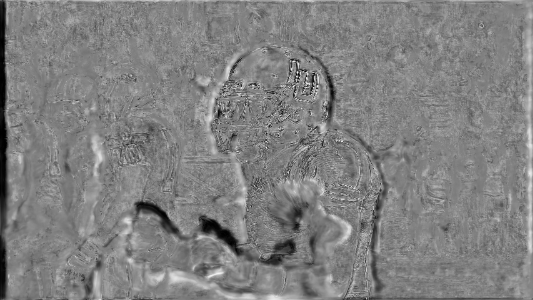} & 
\includegraphics[width=0.43\linewidth]{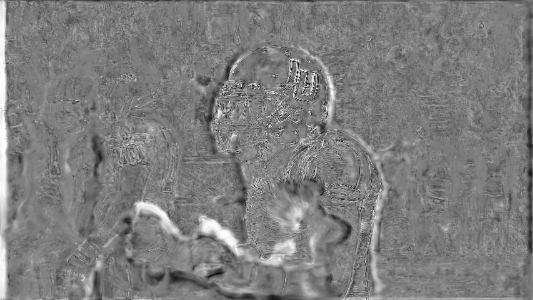} \\
$V_{t\leftarrow0}$ & $V_{t\leftarrow1}$ \\
\includegraphics[width=0.43\linewidth]{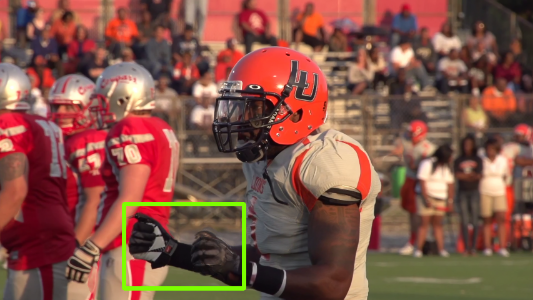} & 
\includegraphics[width=0.43\linewidth]{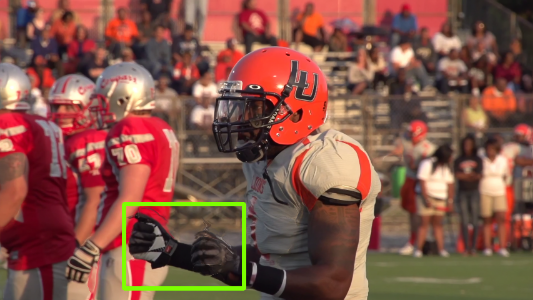} \\
\includegraphics[width=0.43\linewidth]{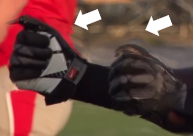} & 
\includegraphics[width=0.43\linewidth]{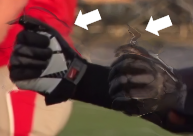} \\
$\hat{I}_t$ & $\hat{I}_t$ w/o visibility maps \\
PSNR=30.23 & PSNR=30.06
\end{tabular}
\caption{Samples of predicted visibility maps (best viewed in color), where $t\!=\!0.5$. The arms move downwards from $T\!=\!0$ to $T\!=\!1$. So the area right above the arm at $T\!=\!0$ is visible at $t$ but the area right above the arm at $T\!=\!1$ is occluded (\ie, invisible) at $t$. The visibility maps in the fourth row clearly show this phenomenon. The white area around arms in $V_{t\leftarrow0}$ indicate such pixels in $I_0$ contribute most to the synthesized $\hat{I}_t$ while the occluded pixels in $I_1$ have little contribution. Similar phenomena also happen around motion boundaries (\eg, around bodies of the athletes).}
\label{fig:vis_maps}
\end{figure} 

As discussed in Section~\ref{subsec:frame_synth}, visibility maps are essential to handle occlusions. Thus, We also predict two visibility maps $V_{t\leftarrow0}$ and $V_{t\leftarrow1}$ using the flow interpolation CNN, and enforce them to satisfy the following constraint
\begin{align}
V_{t\leftarrow0} = 1 - V_{t\leftarrow1}.
\end{align}
Without such a constraint, the network training diverges. Intuitively, $V_{t\leftarrow0}(p)=0$ implies $V_{t\leftarrow1}(p)=1$, meaning that the pixel $p$ from $I_0$ is occluded at $T=t$, we should fully trust $I_1$ and vice versa. Note that it rarely happens that a pixel at time $t$ is occluded both at time $0$ and $1$. Since we use soft visibility maps, when the pixel $p$ is visible both in $I_0$ and $I_1$, the network learns to adaptively combine the information from two images, similarly to the matting effect~\cite{Rhemann09A}. Samples of learned visibility maps are shown in the fourth row of~\figref{fig:vis_maps}.


In order to do flow interpolation, we need to first compute the bi-directional optical flow between the two input images. Recent advances in deep learning for optical flow have demonstrated great potential to leverage deep CNNs to reliably estimate optical flow. In this paper, we train a flow computation CNN, taking two input images $I_0$ and $I_1$, to jointly predict the forward optical flow $F_{0\rightarrow1}$ and backward optical flow $F_{1\rightarrow0}$ between them.

\begin{figure}
\setlength{\belowcaptionskip}{-10pt}
    \centering
    \includegraphics[width=1.05\linewidth]{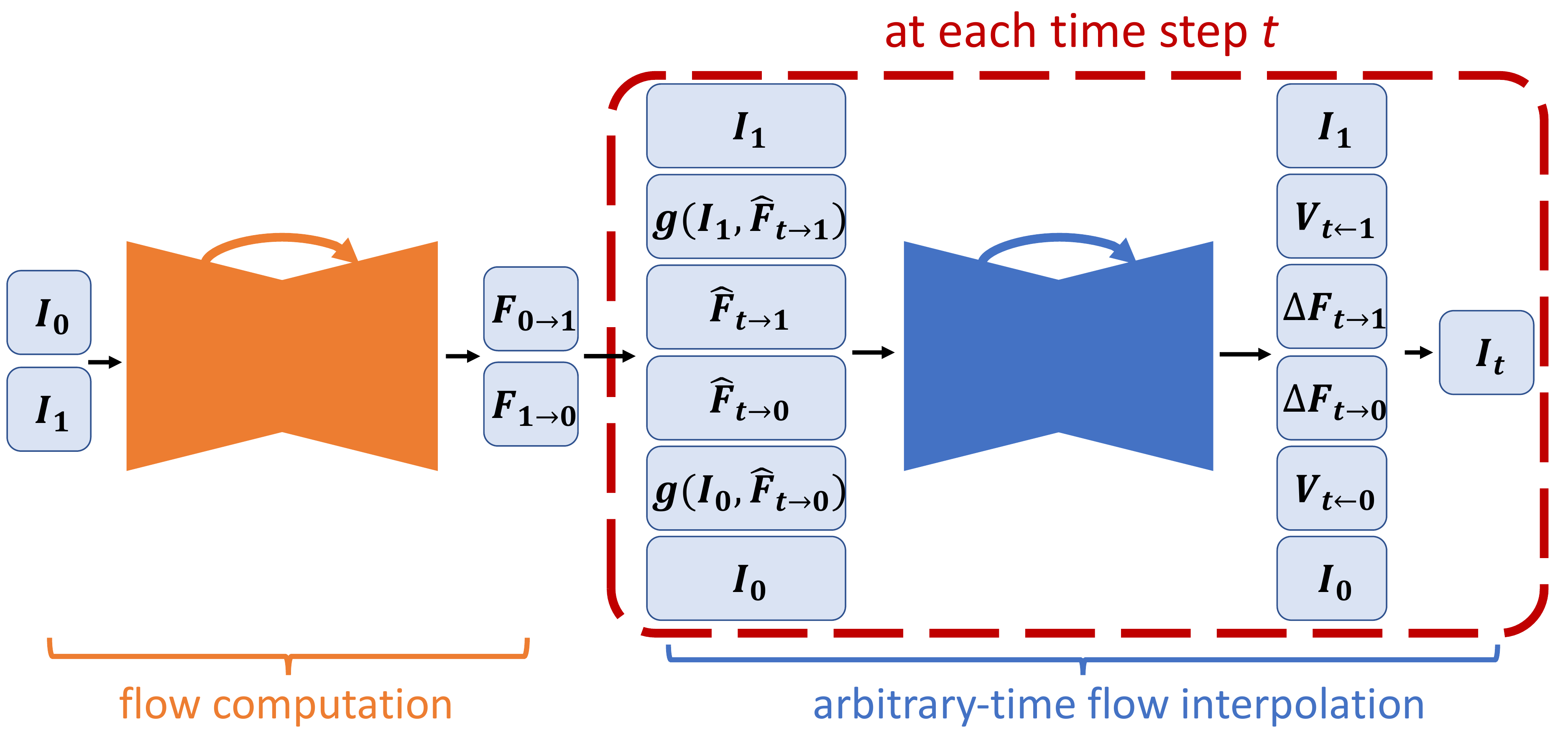}
    \caption{Network architecture of our approach.}
    \label{fig:pipeline}
\end{figure}%

Our entire network is summarized in~\figref{fig:pipeline}. 
For the flow computation and flow interpolation CNNs, we adopt the U-Net architecture~\cite{ronneberger2015u}. The U-Net is a fully convolutional neural network, consisting of an encoder and a decoder, with skip connections between the encoder and decoder features at the same spatial resolution 
For both networks, we have 6 hierarchies in the encoder, consisting of two convolutional and one Leaky ReLU ($\alpha\!=\!0.1$) layers. At the end of each hierarchy except the last one, an average pooling layer with a stride of 2 is used to decrease the spatial dimension. There are 5 hierarchies in the decoder part. At the beginning of each hierarchy, a bilinear upsampling layer is used to increase the spatial dimension by a factor of 2, followed by two convolutional and Leaky ReLU layers. 

For the flow computation CNN, it is crucial to have large filters in the first few layers of the encoder to capture long-range motion. 
We therefore use $7\times7$ kernels in the first two convoluional layers and $5\times5$ in the second hierarchy. For layers in the rest of entire network, we use $3\times 3$ convolutional kernels. The detailed configuration of the network is described in our supplementary material.

We found concatenating output of the encoders in two networks together as input to the decoder of the flow interpolation network yields slightly better results. Moreover, instead of directly predicting the intermediate optical flow in the flow interpolation network, we found it performs slightly better to predict intermediate optical flow residuals. In specific, the flow interpolation network predicts $\Delta F_{t\rightarrow 0}$ and $\Delta F_{t\rightarrow 1}$. We then have
\begin{align}
F_{t\rightarrow 0} = \hat{F}_{t\rightarrow 0} + \Delta F_{t\rightarrow 0} \notag\\
F_{t\rightarrow 1} = \hat{F}_{t\rightarrow 1} + \Delta F_{t\rightarrow 1} 
\end{align}

\subsection{Training}
\label{subsec:training}
Given input images $I_0$ and $I_1$, a set of intermediate frames $\{I_{t_i}\}_{i=1}^N$ between them, where $t_i\in(0, 1)$, and our predictions of intermediate frames $\{\hat{I}_{t_i}\}_{i=1}^N$, our loss function is a linear combination of four terms:
\begin{align}
l = \lambda_r l_r + \lambda_p l_p + \lambda_w l_w + \lambda_s l_s.
\end{align}

\textit{Reconstruction loss} $l_r$ models how good the reconstruction of the intermediate frames is:
\begin{align}
l_r = \frac{1}{N}\sum_{i=1}^N\|\hat{I}_{t_i} - I_{t_i}\|_1.
\end{align}
Such a reconstruction loss is defined in the RGB space, where pixel values are in the range $[0, 255]$.

\textit{Perceptual loss}. Even though we use the $L_1$ loss to model the reconstruction error of intermediate frames, it might still cause blur in the predictions. We therefore use a perceptual loss~\cite{johnson16perceptual} to preserve details of the predictions and make interpolated frames sharper, similar to~\cite{niklaus17video_iccv}. Specifically, the perceptual loss $l_p$ is defined as
\begin{align}
l_p = \frac{1}{N}\sum_{i=1}^N\|\phi(\hat{I}_t) - \phi(I_t)\|_2,
\end{align}
where $\phi$ denote the \texttt{conv4\_3} features of an ImageNet pre-trained VGG16 model~\cite{simonyan14very}

\textit{Warping loss}. Besides intermediate predictions, 
we also introduce the warping loss $l_w$ to model the quality of the computed optical flow, defined as
\begin{align}
l_w = &\|I_0\!-\!g(I_1, F_{0\rightarrow1})\|_1 \!+\! \|I_1\!-\!g(I_0, F_{1\rightarrow0})\|_1 \!+\! \\
	  & \hspace{-8mm} \frac{1}{N}\sum_{i=1}^N\|I_{t_i}\!-\!g(I_0, \hat{F}_{t_i\rightarrow 0})\|_1\!+\!\frac{1}{N}\sum_{i=1}^N\|I_{t_i}\!-\!g(I_1, \hat{F}_{t_i\rightarrow 1})\|_1.\notag
\end{align}

\textit{Smoothness loss}. Finally, we add a smoothness term~\cite{liu17video} to encourage neighboring pixels to have similar flow values:
\begin{align}
l_s = \|\nabla F_{0\rightarrow1}\|_1 + \|\nabla F_{1\rightarrow0}\|_1.
\end{align}
The weights have been set empirically using a validation set as $\lambda_r\!=\!0.8, \lambda_p\!=\!0.005, \lambda_w\!=\!0.4$, and $\lambda_s\!=\!1$. 
Every component of our network is differentiable, including warping and flow approximation. Thus our model can be end-to-end trained. 

\begin{table}[t]
\centering
\renewcommand{\tabcolsep}{1pt}
\caption{Statistics of dataset we use to train our network.}
\label{tab:dataset_statistics}
\begin{tabular}{@{\extracolsep{3pt}}lcc}
\hline
 & Adobe240\fps~\cite{su16deep} & YouTube240\fps \\
\hline
\#video clips & 118 & 1,014 \\
\#video frames & 79,768 & 296,352 \\
mean \#frames per clip & 670.3 & 293.1 \\
resolution & 720p & 720p \\
\hline
\end{tabular}
\end{table}

\begin{figure}[h]
\centering
\setlength{\belowcaptionskip}{-10pt}
\renewcommand{\tabcolsep}{0.5pt}
\begin{tabular}{@{\extracolsep{0.5pt}}ccc}
\includegraphics[width=0.31\linewidth]{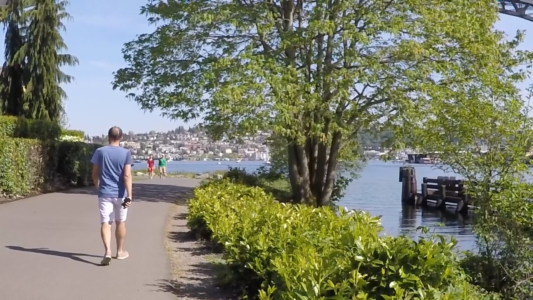} & 
\includegraphics[width=0.31\linewidth]{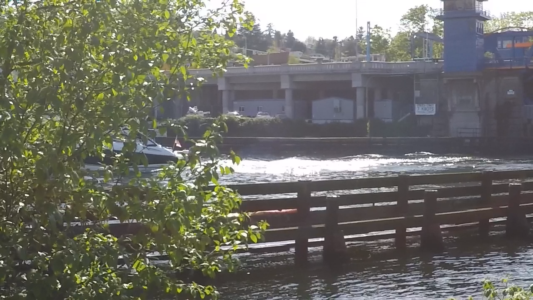} & 
\includegraphics[width=0.31\linewidth]{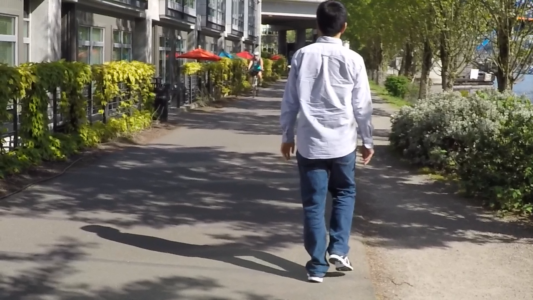} \\
\includegraphics[width=0.31\linewidth]{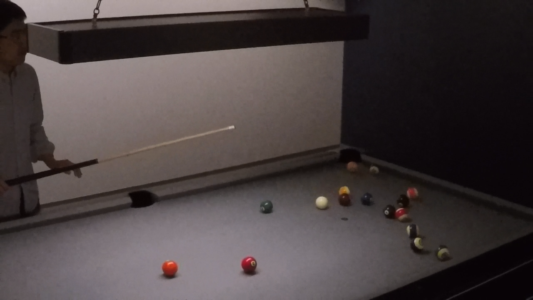} & 
\includegraphics[width=0.31\linewidth]{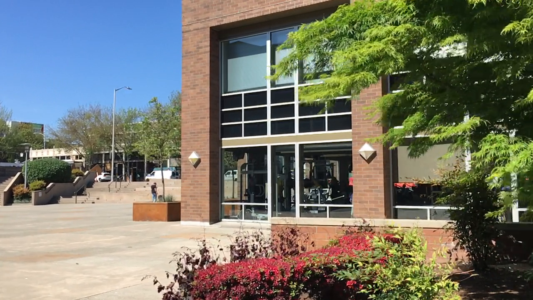} & 
\includegraphics[width=0.31\linewidth]{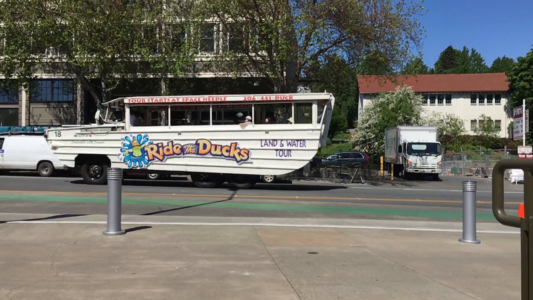} \\
\multicolumn{3}{c}{Adobe240\fps} \\
\includegraphics[width=0.31\linewidth]{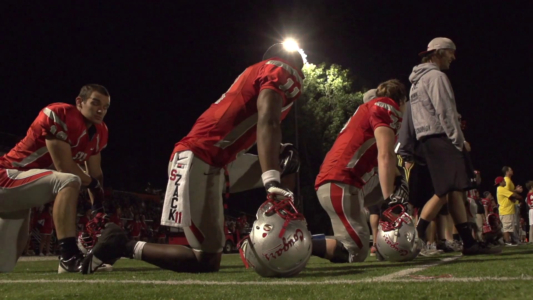} & 
\includegraphics[width=0.31\linewidth]{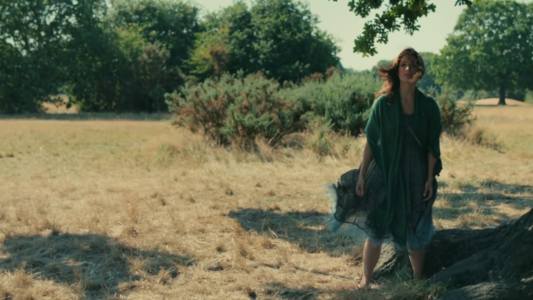} & 
\includegraphics[width=0.31\linewidth]{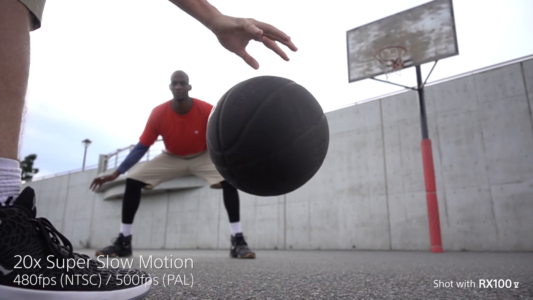} \\
\includegraphics[width=0.31\linewidth]{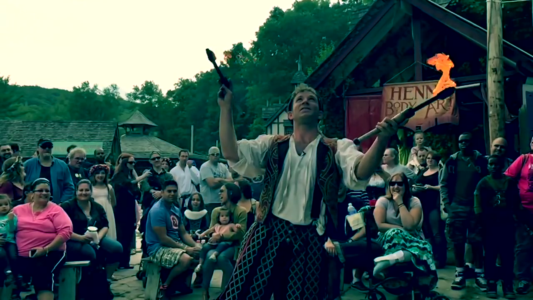} & 
\includegraphics[width=0.31\linewidth]{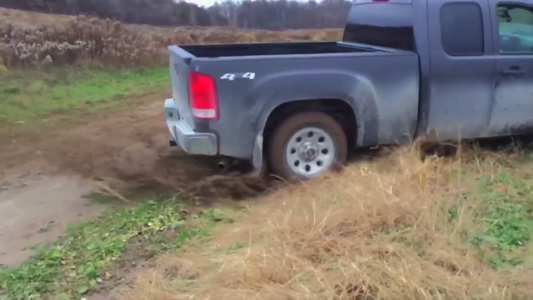} & 
\includegraphics[width=0.31\linewidth]{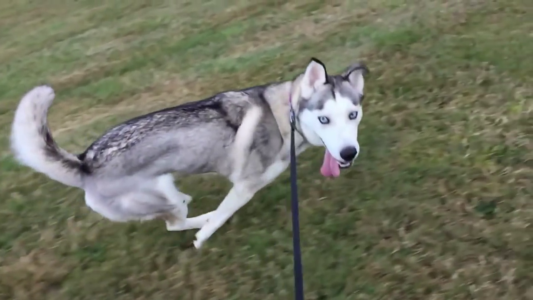} \\
\multicolumn{3}{c}{YouTube240\fps} \\
\end{tabular}
\caption{Snapshot of our training data.}
\label{fig:dataset_snapshot}
\end{figure}

\section{Experiments}

\subsection{Dataset} 
To train our network, we use the 240\fps videos from~\cite{su16deep}, taken with hand-held cameras. We also collect a dataset of 240\fps videos from YouTube. Table~\ref{tab:dataset_statistics} summarizes the statistics of the two datasets and Fig.~\ref{fig:dataset_snapshot} shows a snapshot of randomly sampled video frames. In total, we have 1,132 video clips and 376K individual video frames. There are a great variety of scenes in both datasets, from indoor to outdoor, from static to moving cameras, from daily activities to professional sports, etc. 

We train our network using all of our data and test our model on several independent datasets, including the Middlebury benchmark~\cite{baker11a}, UCF101~\cite{ucf101}, slowflow dataset~\cite{Janai2017CVPR}, and high-frame-rate Sintel sequences~\cite{Janai2017CVPR}. For Middlebury, we submit our single-frame video interpolation results of eight sequences to its evaluation server. For UCF101, in every triple of frames, the first and third ones are used as input to predict the second frame using 379 sequences provided by~\cite{liu17video}. The slowflow dataset contains 46 videos taken with professional high-speed cameras. We use the first and eighth video frames as input, and interpolate intermediate 7 frames, equivalent to converting a 30\fps video to a 240\fps one. The original Sintel sequences~\cite{Butler:ECCV:2012} were rendered at 24 fps. 13 of them were re-rendered at 1008 fps~\cite{Janai2017CVPR}. To convert from 24\fps to 1008\fps using a video frame interpolation approach, one needs to insert 41 in-between frames. However, as discussed in the introduction, it is not directly possible with recursive single-frame interpolation methods~\cite{niklaus17video_cvpr,niklaus17video_iccv,liu17video} to do so. Therefore, we instead predict 31 in-between frames for fair comparisons with previous methods.

Our network is trained using the Adam optimizer~\cite{kingma2014adam} for 500 epochs. The learning rate is initialized to be 0.0001 and  decreased by a factor of 10 every 200 epochs. During training, all video clips are first divided into shorter ones with 12 frames in each and there is no overlap between any of two clips. For data augmentation, we randomly reverse the direction of entire sequence and select 9 consecutive frames for training. On the image level, each video frame is resized to have a shorter spatial dimension of 360 and a random crop of $352\times 352$ plus horizontal flip are performed.

For evaluation, we report Peak Signal-to-Noise Ratio (PSNR) and Structural Similarity Index (SSIM) scores between predictions and ground-truth in-between video frames, 
as well as the interpolation error (IE)~\cite{baker11a}, which is defined as root-mean-squared (RMS) difference between the ground-truth image and the interpolated image.

\begin{table}
\caption{Effectiveness of multi-frame video interpolation on the \emph{Adobe240-fps} dataset.}
\label{tab:ablation_multi_interp}
\centering
\renewcommand{\tabcolsep}{5pt}
\begin{tabular}{@{\extracolsep{5pt}}cccc}
\toprule
 & PSNR & SSIM & IE \\
\midrule
1 interp & 30.26 & 0.909 & 8.85 \\
3 interp & 31.02 & 0.917 & 8.43 \\
7 interp & \textbf{31.19} & \textbf{0.918} & \textbf{8.30} \\
\bottomrule
\end{tabular}
\end{table}

\begin{table}
\caption{Effectiveness of different components of our model on the \emph{Adobe240-fps} dataset.}
\label{tab:ablation_components}
\centering
\renewcommand{\tabcolsep}{3.5pt}
\begin{tabular}{@{\extracolsep{4pt}}lccc}
\toprule
 & PSNR & SSIM & IE \\
\midrule
w/o flow interpolation & 30.34 & 0.908 & 8.93 \\
w/o vis map & 31.16 & 0.918 & 8.33 \\
w/o perceptual loss & 30.96 & 0.916 & 8.50 \\
w/o warping loss & 30.52 & 0.910 & 8.80 \\
w/o smoothness loss & \textbf{31.19} & \textbf{0.918} & \textbf{8.26} \\
\midrule
full model & \textbf{31.19} & \textbf{0.918} & 8.30 \\
\bottomrule
\end{tabular}

\end{table}

\subsection{Ablation Studies}
In this section, we perform ablation studies to analyze our model. For the first two experiments, we randomly sampled 107 videos from Adobe240\fps dataset for training and the remaining 12 ones for testing. 


\noindent\textbf{Effectiveness of multi-frame video interpolation.} 
We first test whether jointly predicting several in-between frames improves the video interpolation results. Intuitively, predicting a set of in-between frames together might implicitly enforce the network to generate temporally coherent sequences. 


To this end, we train three variants of our model: predicting intermediate single, three, and seven frames, which are all evenly distributed across time steps. At test time, we use each model to predict seven in-between frames. Table~\ref{tab:ablation_multi_interp} clearly demonstrates that the more intermediate frames we predict during training, the better the model is. 

\begin{figure}
\centering
\begin{tabular}{c}
\includegraphics[width=0.92\linewidth]{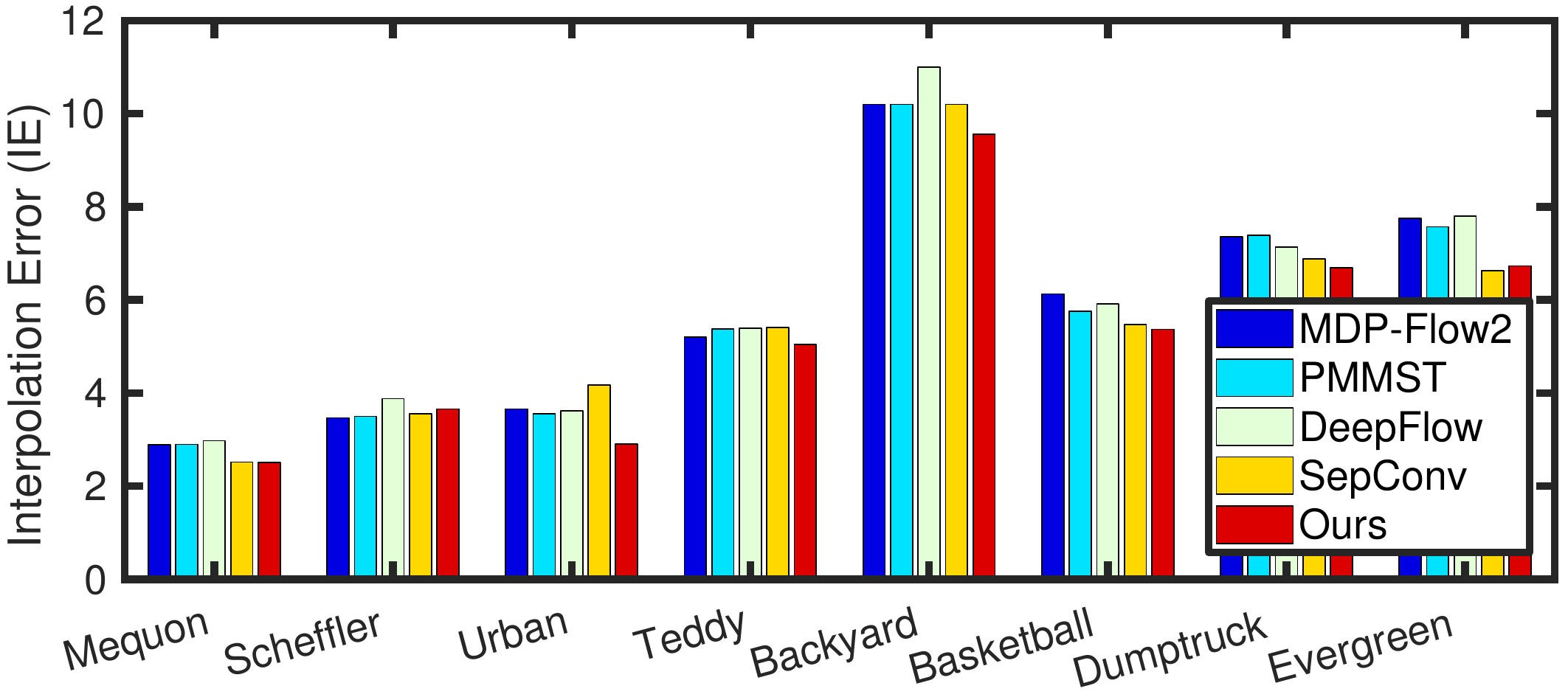} \\
\end{tabular}
\caption{Performance comparisons on each sequence of the Middlebury dataset. Numbers are obtained from the Middlebury evaluation server.}
\label{fig:middlebury}
\end{figure}

\noindent\textbf{Impact of different components design.} We also investigate the contribution of each component in our model. In particular, we study the impact of flow interpolation by removing the flow refinement from the second U-Net (but keep using the visibility maps). We further study the use of visibility maps as means of occlusion reasoning. We can observe from Table~\ref{tab:ablation_components} that removing each of three components harms performance. Particularly, the flow interpolation plays a crucial role, which verifies our motivation to introduce the second learned network to refine intermediate optical flow approximations. Adding visibility maps improves the interpolation performance slightly. Without it, there are artifacts generated around motion boundaries, as shown in Figure~\ref{fig:vis_maps}. Both of these validate our hypothesis that jointly learning motion interpretation and occlusion reasoning helps video interpolation.

We also study different loss terms, where the warping loss is the most important one. Although adding the smoothness terms slightly hurts the performance quantitatively, we fount it is useful to generate visually appealing optical flow between input images.

\begin{table}
\setlength{\belowcaptionskip}{-10pt}
\caption{Results on the \emph{UCF101} dataset.}
\label{tab:quant_comp_ucf101}
\centering
\begin{tabular}{lccc}
\toprule
 & PSNR & SSIM & IE \\
\midrule
Phase-Based~\cite{meyer15phase} & 32.35 & 0.924 & 8.84 \\
FlowNet2~\cite{baker11a,ilg16flownet2} & 32.30 & 0.930 & 8.40 \\
DVF~\cite{liu17video} & 32.46 & 0.930 & 8.27 \\
SepConv~\cite{niklaus17video_iccv} & 33.02 & 0.935 & 8.03 \\
\midrule
Ours (Adobe240\fps) & 32.84 & 0.935 & 8.04 \\
Ours & \textbf{33.14} & \textbf{0.938} & \textbf{7.80} \\
\bottomrule
\end{tabular}
\end{table}


\noindent\textbf{Impact of the number of training samples.} Finally, we investigate the effect of the number of training samples. We compare two models: one trained on the Adobe240\fps dataset only and the other one trained on our full dataset. The performance of these two models on the UCF101 dataset can be found in last two rows Table~\ref{tab:quant_comp_ucf101}. We can see that our model benefits from more training data.

\subsection{Comparison with state-of-the-art methods} 
In this section, we compare our approach with state-of-the-art methods including phase-based interpolation~\cite{meyer15phase}, separable adaptive convolution (SepConv)~\cite{niklaus17video_iccv}, and deep voxel flow (DVF)~\cite{liu17video}. We also implement a baseline approach using the interpolation algorithm presented in~\cite{baker11a}, where we use FlowNet2~\cite{ilg16flownet2} to compute the bi-directional optical flow results between two input images. FlowNet2 is good at capturing global background motion and recovering sharp motion boundaries for the optical flow. Thus, when coupled with occlusion reasoning~\cite{baker11a}, FlowNet2 serves as a strong baseline.

\begin{table}
\caption{Results on the \emph{slowflow} dataset.}
\label{tab:quant_comp_slowflow}
\centering
\begin{tabular}{lccc}
\toprule
 & PSNR & SSIM & IE \\
\midrule
Phase-Based~\cite{meyer15phase} & 31.05 & 0.858 & 8.21 \\
FlowNet2~\cite{baker11a,ilg16flownet2} & 34.06 & \textbf{0.924} & \textbf{5.35} \\
SepConv~\cite{niklaus17video_iccv} & 32.69 & 0.893 & 6.79 \\
\midrule
Ours & \textbf{34.19} & \textbf{0.924} & 6.14 \\
\bottomrule
\end{tabular}
\end{table}

\begin{table}
\setlength{\belowcaptionskip}{-5pt}
\caption{Results on the high-frame-rate \emph{Sintel} dataset.}
\label{tab:quant_comp_hfr_sintel}
\centering
\begin{tabular}{lccc}
\toprule
 & PSNR & SSIM & IE \\
\midrule
Phase-Based~\cite{meyer15phase} & 28.67 & 0.840 & 10.24 \\
FlowNet2~\cite{baker11a,ilg16flownet2} & 30.79 & 0.922 & 5.78 \\
SepConv~\cite{niklaus17video_iccv} & 31.51 & 0.911 & 6.61 \\
\midrule
Ours & \textbf{32.38} & \textbf{0.927} & \textbf{5.42} \\
\bottomrule
\end{tabular}
\end{table}

\begin{figure*}
\centering
\renewcommand{\tabcolsep}{0.5pt}
\begin{tabular}{ccccccc}
\includegraphics[width=0.13\linewidth]{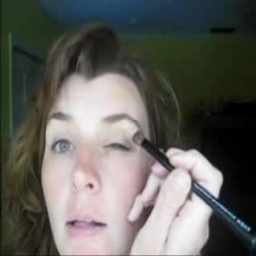} & 
\includegraphics[width=0.13\linewidth]{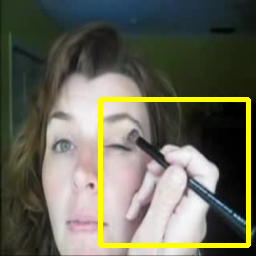} & 
\includegraphics[width=0.13\linewidth]{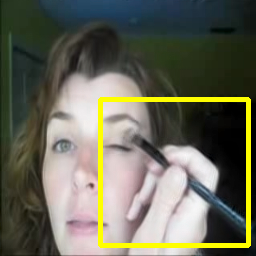} & 
\includegraphics[width=0.13\linewidth]{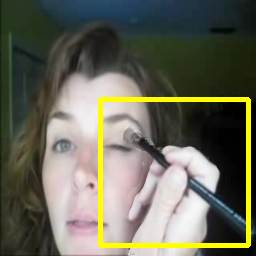} & 
\includegraphics[width=0.13\linewidth]{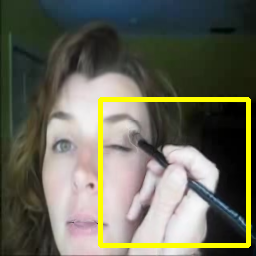} & 
\includegraphics[width=0.13\linewidth]{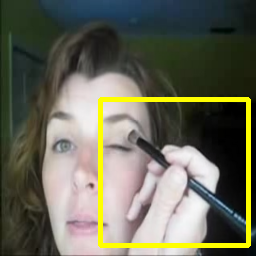} & 
\includegraphics[width=0.13\linewidth]{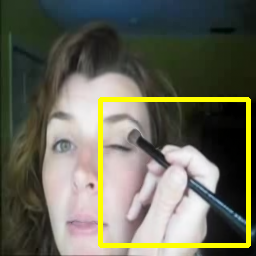} \\
\includegraphics[width=0.13\linewidth]{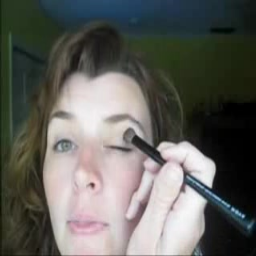} & 
\includegraphics[width=0.13\linewidth]{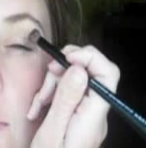} & 
\includegraphics[width=0.13\linewidth]{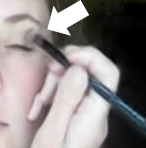} & 
\includegraphics[width=0.13\linewidth]{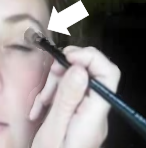} & 
\includegraphics[width=0.13\linewidth]{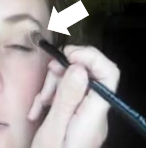} & 
\includegraphics[width=0.13\linewidth]{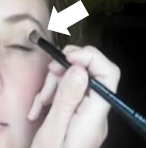} & 
\includegraphics[width=0.13\linewidth]{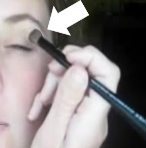} \\
\small 2 input images & \small  actual in-between & \small PhaseBased~\cite{meyer15phase} & \small FlowNet2 \cite{baker11a,ilg16flownet2} & \small DVF \cite{liu17video} & \small SepConv \cite{niklaus17video_iccv} & \small ours
\end{tabular}
\caption{Visual results of a sample from UCF101. Our model produces less artifacts around the brush and the hand (best viewed in color). Please see the supplementary material for more image and \emph{video} results. }
\label{fig:interpolation_results}
\end{figure*}

\noindent\textbf{Single-frame video interpolation.} The interpolation error (IE) scores on each sequence form the Middlebury dataset are shown in Figure~\ref{fig:middlebury}. In addition to SepConv, we also compare our model with other three top-performing models on the Middlebury dataset\footnote{\url{http://vision.middlebury.edu/flow/eval/results/results-i1.php}}, where the interpolation algorithm~\cite{baker11a} is coupled with different optical flow methods including MDP-Flow2~\cite{xu2012motion}, PMMST~\cite{pmmst}, and DeepFlow~\cite{deepflow}. Our model achieves the best performance on 6 out of all 8 sequences. Particularly, the \texttt{Urban} sequence is generated synthetically and the \texttt{Teddy} sequence contains actually two stereo pairs. The performance of our model validates the generalization ability of our approach.

On UCF101, we compute all metrics using the motion masks provided by~\cite{liu17video}. The quantitative results are shown in Table~\ref{tab:quant_comp_ucf101}, highlighting the performance of each interpolation model's capacity to deal with challenging motion regions. Our model consistently outperforms both non-neural~\cite{meyer15phase} and CNN-based approaches~\cite{niklaus17video_iccv,liu17video}. Sample interpolation results on a sample from UCF101 can be found at Figure~\ref{fig:interpolation_results}. More results can be found in our supplementary material.

\noindent\textbf{Multi-frame video interpolation.} For the slowflow dataset, we predict 7 in-between frames. All experiments are performed on the half-resolution images with a spatial dimension of $1280\times1024$. On this dataset, our approach achieves the best PSNR and SSIM scores and FlowNet2 achieves the best SSIM and $L_1$ error scores. FlowNet2 is good at capturing global motions and thus produces sharp prediction results on those background regions, which follow a global motion pattern. Detailed visual comparisons can be found in our supplementary material.

\begin{figure}
\centering
\setlength{\belowcaptionskip}{-10pt}
\includegraphics[width=0.8\linewidth]{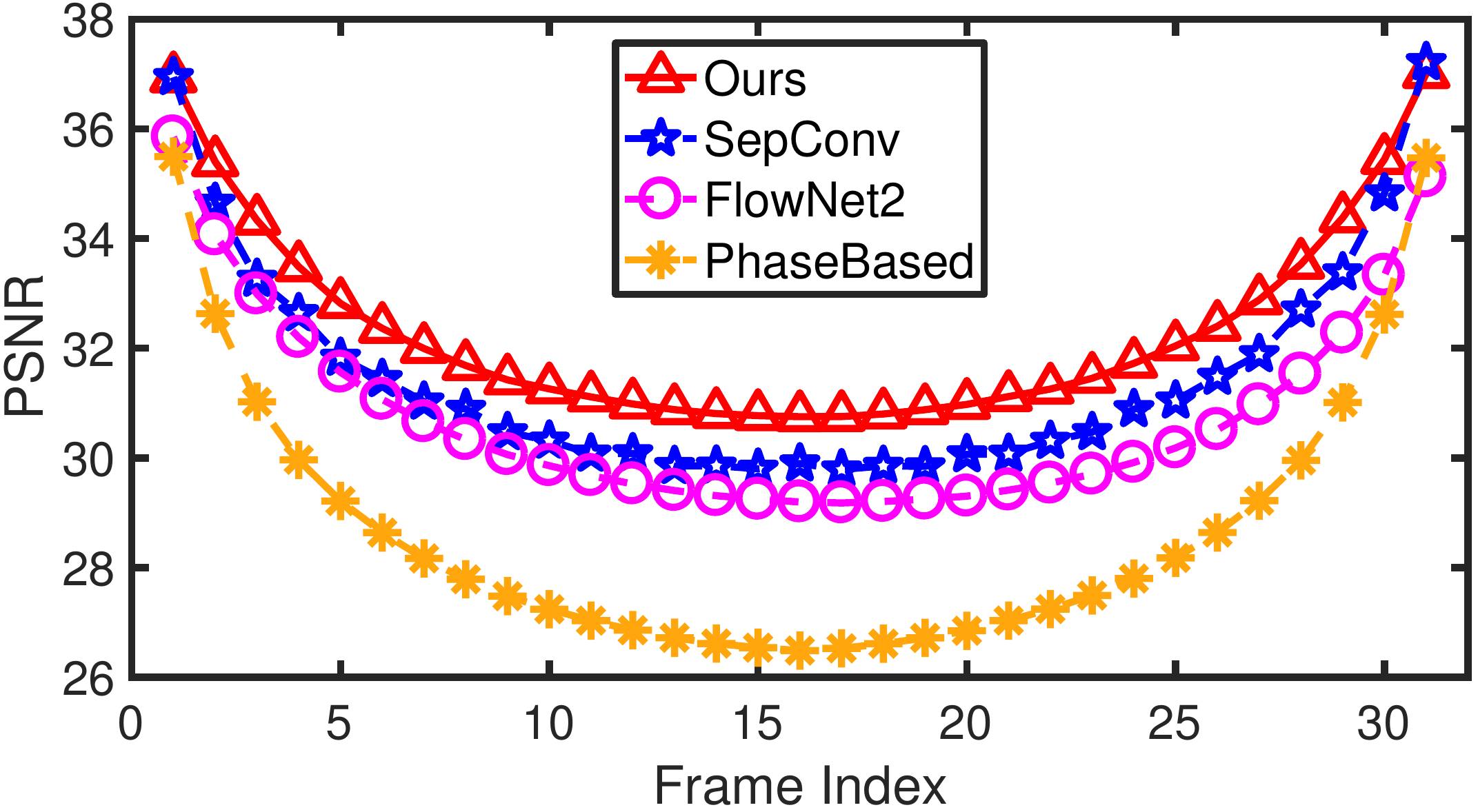}
\caption{PSNR at each time step when generating 31 intermediate frames on the high-frame-rate Sintel dataset.}
\label{fig:psnr_hfr_sintel}
\end{figure}

On the challenging high-frame-rate Sintel dataset, our approach significantly outperforms all other methods. We also show the PSNR scores at each time step in Figure~\ref{fig:psnr_hfr_sintel}. Our approach produces the best predictions for each in-between time step except slightly worse than SepConv at the last time step.


In summary, our approach achieves state-of-the-art results over all datasets, generating single or multiple intermediate frames. It is remarkable, considering the fact our model can be directly applied to different scenarios without any modification. 

\subsection{Unsupervised Optical Flow}
Our video frame interpolation approach has an unsupervised (self-supervised) network (the flow computation CNN) that can compute the bi-directional optical flow between two input images. Following~\cite{liu17video}, we evaluate our unsupervised forward optical flow results on the testing set of KITTI 2012 optical flow benchmark~\cite{Geiger:2012:KITTI}. The average end-point error (EPE) scores of different methods are reported in Table~\ref{tab:optical_flow_kitti_2012}. Compared with previous unsupervised method DVF~\cite{liu17video}, our model achieves an average EPE of 13.0, an 11\% relative improvement. Very likely this improvement results from the multi-frame video interpolation setting, as DVF~\cite{liu17video} has a similar U-Net architecture to ours.


\begin{table}
\centering
\setlength{\belowcaptionskip}{-10pt}
\renewcommand{\tabcolsep}{3.5pt}
\caption{Optical flow results on the KITTI 2012 benchmark. 
}
\label{tab:optical_flow_kitti_2012}
\begin{tabular}{cccccc}
\hline
 & {\small LDOF\cite{Brox:LDOF:2011}} & {\small EpicFlow\cite{EpicFlow}} & {\small FlowNetS\cite{Dosovitskiy:2015Flownet}} & {\small DVF\cite{liu17video}} & {\small Ours} \\
\hline
EPE & 12.4 & 3.8 &  9.1 & 14.6 & 13.0 \\
\hline
\end{tabular}
\end{table}

\section{Conclusion}
We have proposed an end-to-end trainable CNN that can produce as many intermediate video frames as needed between two input images. We first use a flow computation CNN to estimate the bidirectional optical flow between the two input frames, and the two flow fields are linearly fused to approximate the intermediate optical flow fields. We then use a flow interpolation CNN to refine the approximated flow fields and predict soft visibility maps for interpolation. We use more than 1.1K 240\fps video clips to train our network to predict seven intermediate frames. Ablation studies on separate validation sets demonstrate the benefit of flow interpolation and visibility map. Our multi-frame approach consistently outperforms state-of-the-art single frame methods on the Middlebury, UCF101, slowflow, and high-frame-rate Sintel datasets.  For the unsupervised learning of optical flow, our network outperforms the recent DVF method~\cite{liu17video} on the KITTI 2012 benchmark. 







\section*{Acknowledgement}
We would like to thank Oliver Wang for generously sharing the Adobe 240-fps data~\cite{su16deep}.
Yang acknowledges support from NSF (Grant No. 1149783).

\begin{figure*}
\centering
\includegraphics[width=0.8\linewidth]{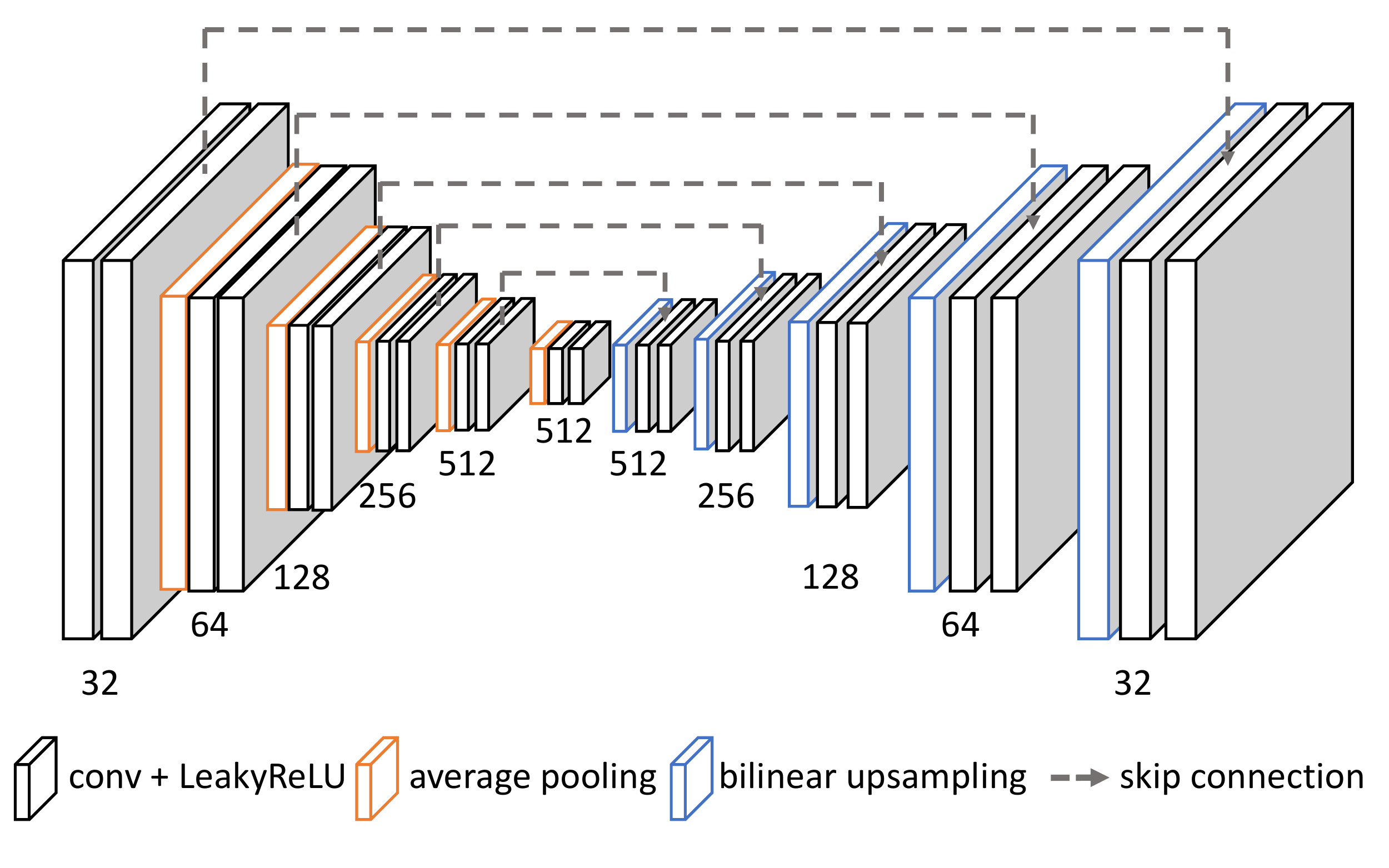} \\
\caption{Illustration of the architecture of our flow computation and flow interpolation CNNs.}
\label{fig:network_architecture}
\end{figure*}

\appendix
\section{Network Architecture}
Our flow computation and flow interpolation CNNs share a similar U-Net architecture, shown in Figure~\ref{fig:network_architecture}.

\section{Visual Comparisons on UCF101}
\begin{figure*}
\centering
\includegraphics[width=0.85\linewidth]{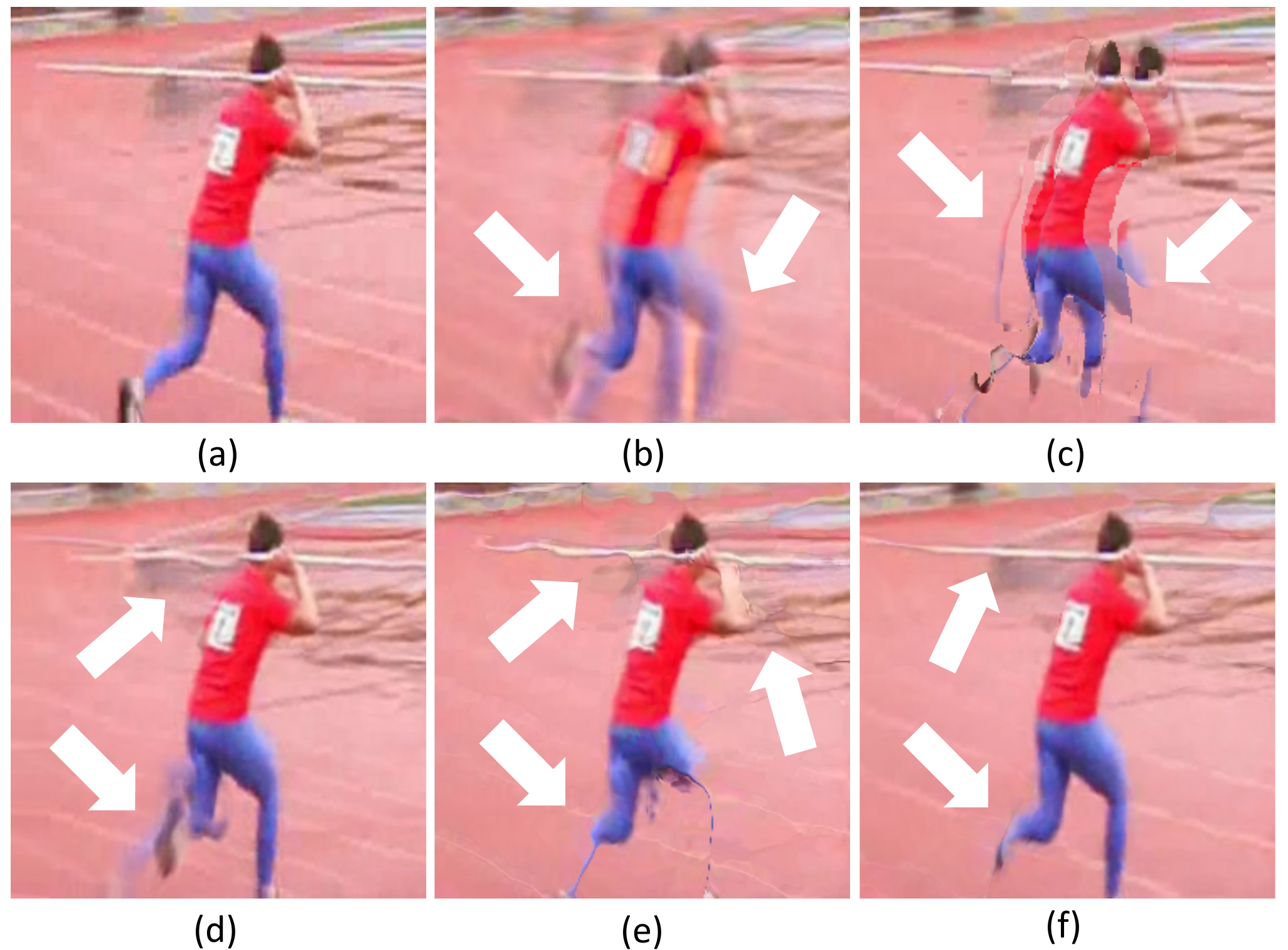} \\
\includegraphics[width=0.85\linewidth]{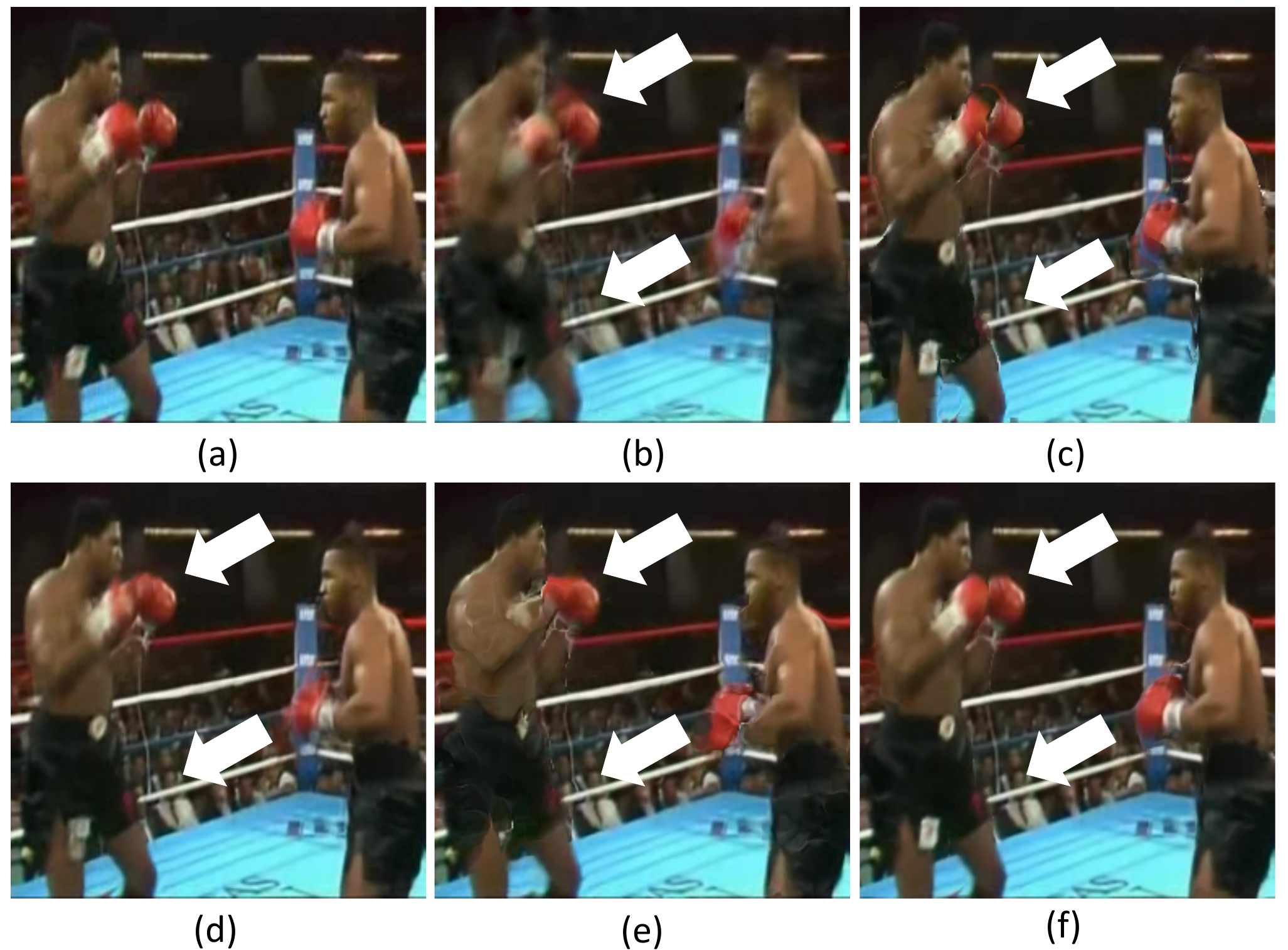}
\caption{Visual comparisons on the UCF101 dataset. (a) actual in-between frame, interpolation results from (b) PhaseBased~\cite{meyer15phase}, (c) FlowNet2~\cite{Baker2009OcclusionInterpolation,ilg16flownet2}, (d) SepConv~\cite{niklaus17video_iccv}, (e) DVF~\cite{liu17video}, and (f) Ours.}
\label{fig:vis_comp_ucf101_1}
\end{figure*}

\begin{figure*}
\centering`'
\includegraphics[width=0.85\linewidth]{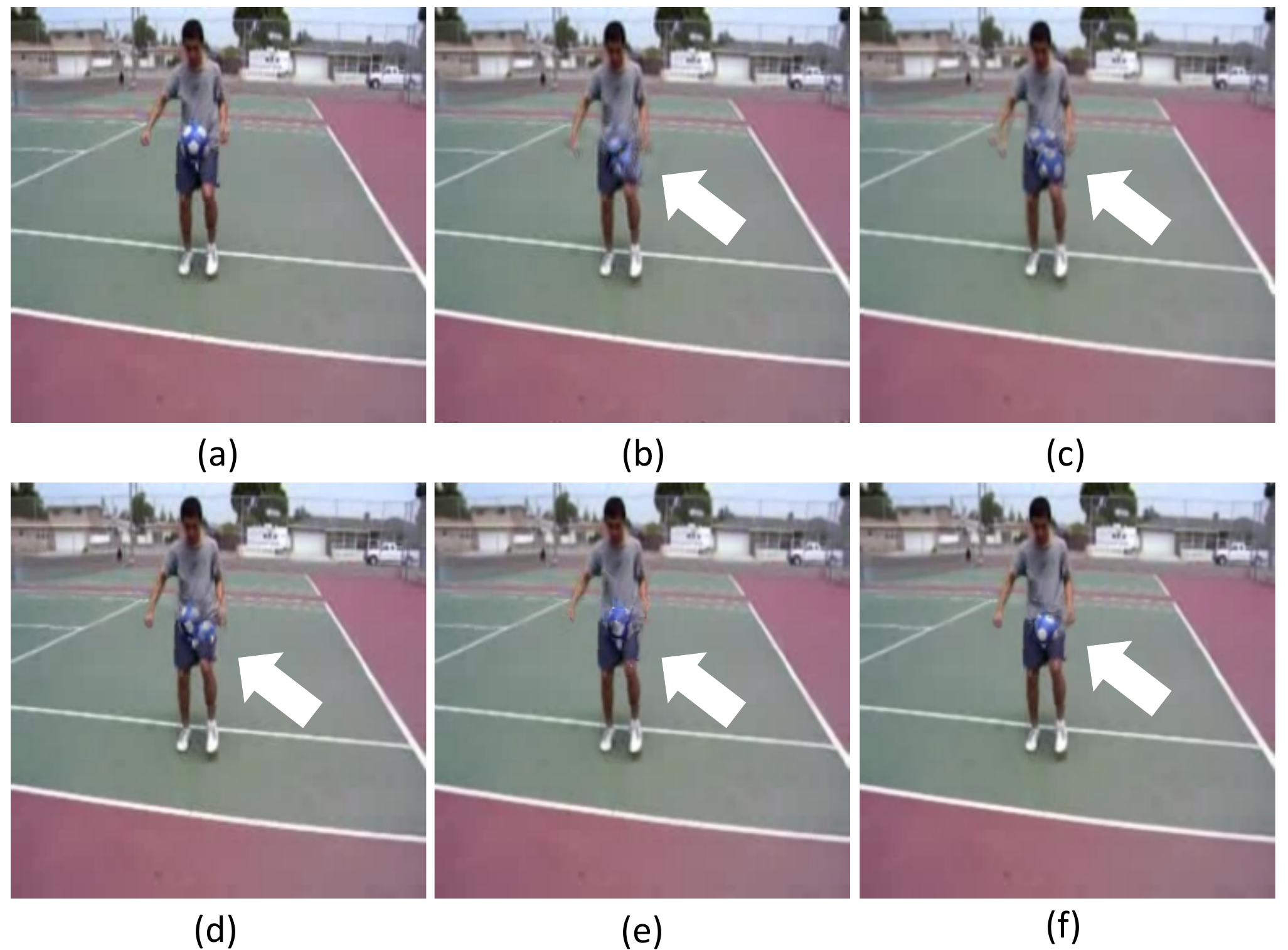} \\
\includegraphics[width=0.85\linewidth]{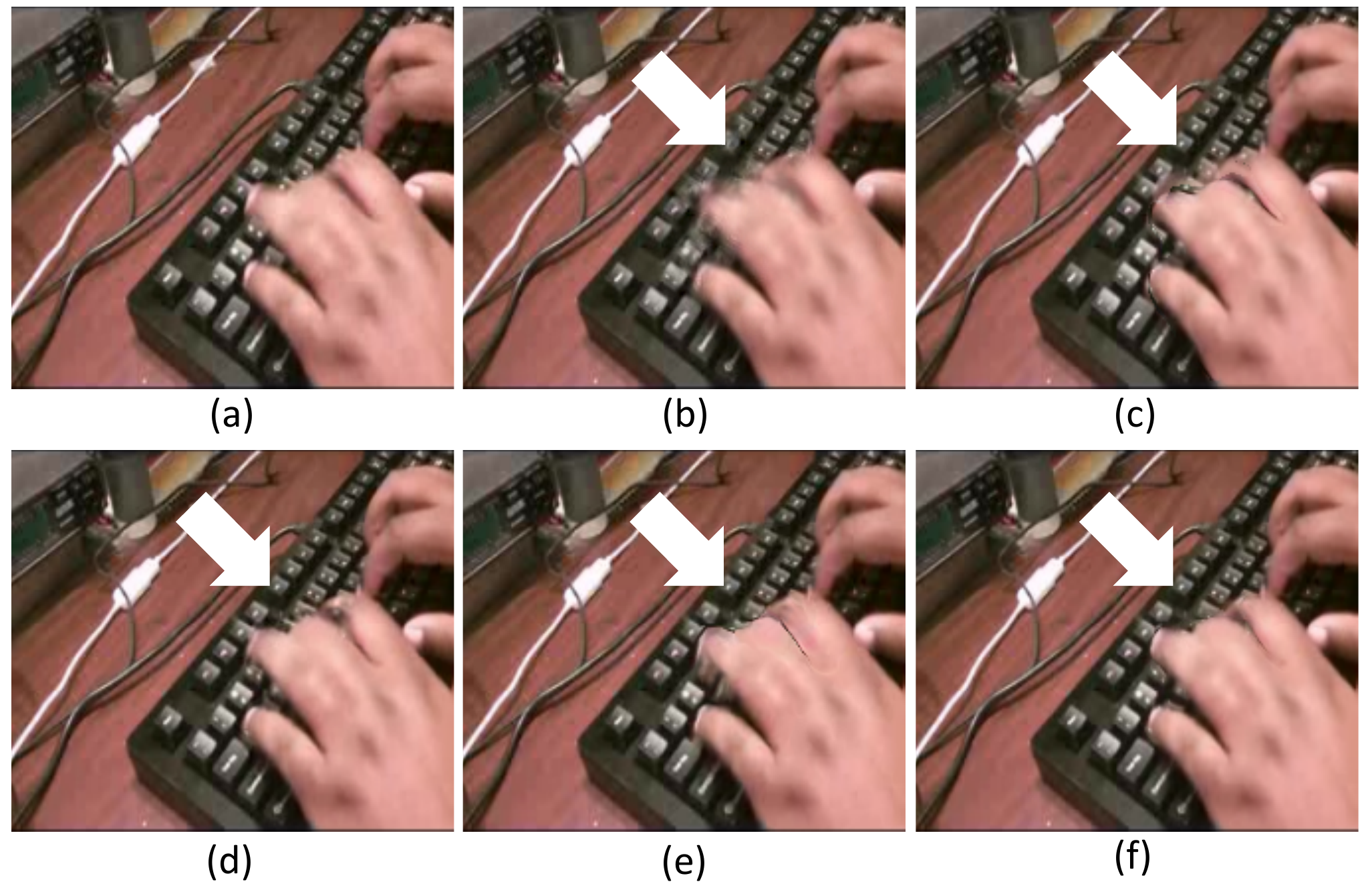}
\caption{Visual comparisons on the UCF101 dataset. (a) actual in-between frame, interpolation results from (b) PhaseBased~\cite{meyer15phase}, (c) FlowNet2~\cite{Baker2009OcclusionInterpolation,ilg16flownet2}, (d) SepConv~\cite{niklaus17video_iccv}, (e) DVF~\cite{liu17video}, and (f) Ours.}
\label{fig:vis_comp_ucf101_2}
\end{figure*}

Figure~\ref{fig:vis_comp_ucf101_1} and Figure~\ref{fig:vis_comp_ucf101_2} show visual comparisons of single-frame interpolation results on the UCF101 dataset. For more visual comparisons, please refer to our supplementary video~\url{http://jianghz.me/projects/superslomo/superslomo_public.mp4}.

{\small
\bibliographystyle{ieee}
\bibliography{egbib}
}

\end{document}